\documentclass[runningheads]{llncs}
\usepackage[T1]{fontenc}
\usepackage{graphicx}
\usepackage{cite}
\usepackage{amsmath,amsfonts}
\usepackage{url}
\usepackage{verbatim}
\usepackage{multirow}
\begin{document}
\title{Radiogenomic Bipartite Graph Representation Learning for Alzheimer's Disease Detection}
%
%
%
\author{Aditya Raj\inst{1} \and
Golrokh Mirzaei\inst{2}}
\authorrunning{Raj et al.}
%
\institute{Department of Electrical and Computer Engineering, Ohio State University, Columbus OH 43210, USA \and
Department of Computer Science and Engineering, Ohio State University, Columbus, OH 43210, USA\\
\email{mirzaei.4@osu.edu}}
\maketitle              

\begin{abstract}
Imaging and genomic data offer distinct and rich features, and their integration can unveil new insights into the complex landscape of diseases. In this study, we present a novel approach utilizing radiogenomic data including structural MRI images and gene expression data, for Alzheimer’s disease detection. Our framework introduces a novel heterogeneous bipartite graph representation learning featuring two distinct node types: genes and images. The network can effectively classify Alzheimer's disease (AD) into three distinct stages:AD, Mild Cognitive Impairment (MCI), and Cognitive Normal (CN) classes, utilizing a small dataset. Additionally, it identified which genes play a significant role in each of these classification groups. We evaluate the performance of our approach using metrics including classification accuracy, recall, precision, and F1 score. The proposed technique holds potential for extending to radiogenomic-based classification to other diseases.


\end{abstract}

\section{Introduction}
Alzheimer's disease (AD) is one of the most prevalent neurodegenerative conditions that progressively impairs cognitive function and overall functionality in individuals. It primarily targets the aging population and is characterized by the accumulation of abnormal protein aggregates in the brain, which leads to the gradual destruction of nerve cells and neural connections. Consequently, those afflicted with Alzheimer's often experience memory loss, difficulties in thinking, problem-solving, and language skills, as well as changes in behavior and personality.

Currently, imaging is one of the common techniques used to assess Alzheimer's disease. Neuroimaging methods, such as MRI and PET scans, offer detailed insights into the brain's structure and function. The benefits of utilizing imaging data as a non-invasive diagnostic tool are significant; however, there are also some limitations associated with this approach. Imaging techniques can be expensive and, on their own, may not provide conclusive results. Additionally, accessibility can be an issue for some individuals, and the process can be time-consuming, often requiring multiple scans for accurate pathology detection. Furthermore, when it comes to structural MRI (sMRI)-based Alzheimer's detection, brain atrophy is not disease-specific and can result from various other conditions \cite{van2021imaging,johnson2012brain}. For example, the degradation of hippocampus volume, considered a vital biomarker for Alzheimer's, has also been associated with several other neurodegenerative disorders, including Parkinson's disease \cite{camicioli2003parkinson}, Huntington's disease \cite{rosas2003evidence}, and even non-neurological conditions like cardiac arrest \cite{fujioka2000hippocampal} and alcohol dependence \cite{agartz1999hippocampal}. Therefore, it becomes essential to develop diagnostic tools that incorporate more than just imaging data. Mirzaei et al. provided  comprehensive surveys of different imaging and machine learning techniques for AD detection\cite{mirzaei2016ImagingMachineLearning}\cite{mirzaei2022MachineLearning}. 

Studies has demonstrated that incorporating imaging and genomic features leads to superior model performance compared to relying solely on either imaging or genomics data in disease detection \cite{salvatore2021radiomics, li2020radiomics}. Genomics can offer supplementary information and often serves as a cost-effective alternative in various cases compared to repeated imaging scans. Thus, the integration of genomics and imaging features holds the promise of designing diagnostic tools that are not only cost-efficient but also capable of providing conclusive assessments. Imaging genomics has recently received attention in AD detection\cite{chaddad2018deep}, AD biomarker prediction \cite{zhou2019dual}, and the progression analysis of Mild Cognitive Impairment (MCIs) to AD \cite{jiang2022using}. 

Majority of current radiogenomic studies integrate imaging features and genomics using standard concatenation in a form of matrix or fusion layer \cite{maddalena2023integrating} \cite{Gupta2019Prediction}. However, this approach may disregard the inherent complexities and interactions within the data.  Other studies have focused on using one type of modality, either genomics or imaging for AD detection. For instance, Mirzaei \cite{Mirzaei2021Ensemble} and Kavitha et al. \cite{Kavitha2022Early} utilized MRI imaging, while Nativio et al. \cite{Nativio2020An} solely focused on genomics. 

In this study, we propose a bipartite graph representation learning (BGRL) for AD detection using radiogenomic data, which includes structural MRI images and gene expression data for three AD driver genes (PSEN1, PSEN2, and APOE). These genes are chosen for their well-known association with Alzheimer’s disease. The bipartite graph is constructed based on two sets of nodes: images and genes. We assume that there are no edges between the nodes within the same modality. Subsequently, a dynamic adjacency matrix is learned through training, which establishes the connections between nodes (image and genes). We introduce a novel aggregation function for learning weights, derived from a randomly sampled prior.  The proposed framework is evaluated using performance metrics, including accuracy, F1 score, recall, and precision.  

\section{Methods}
Given a multimodal dataset consisting pairs of MRI images and gene expression data for three driver genes associated with AD (PSEN1, PSEN2, and APOE), our method learns the structural information of the graph for the downstream classification task. As shown in Fig. 1, the overall framework consists of three parts: 1) 3D autoEncoder denoising for feature extraction, 2) Bipartite graph construction by fusion of imaging and genomics data, and 3) Radiogenomic graph structure learning and classification. In the subsequent sections, we will introduce each part of our proposed framework in detail. 
\begin{figure*}[tb]
\centering
\includegraphics[width=\textwidth]{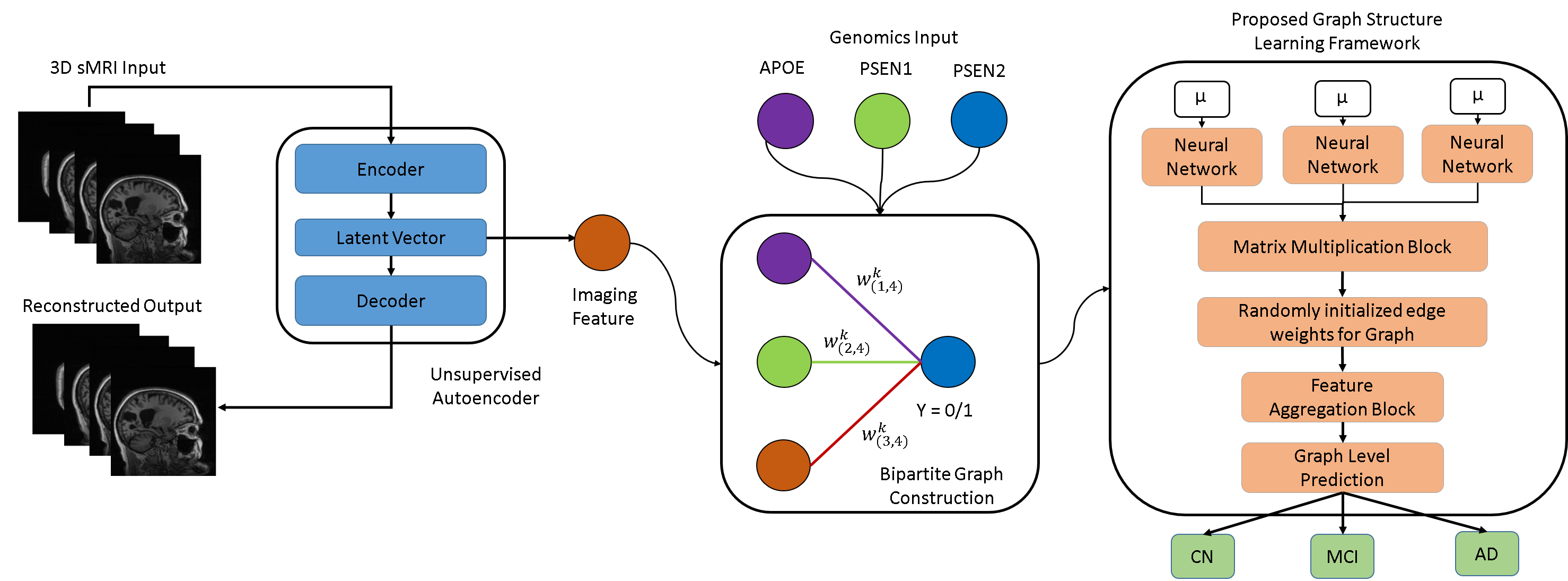}
\caption{\label{fig:new_frame} The proposed framework of bipartite graph representation learning with imaging genomics data.}
\end{figure*}

\subsection{3D Denoising Autoencoder for Feature Extraction}
We utilize a 3D denoising autoencoder to extract features from MRI data. The autoencoder consists of an encoder function $f_{e}$ and a decoder function $f_{d}$. Gaussian noise $\mu \sim N(0,1)$ is added to the normalized feature vector $x$, resulting in a noisy input $x' = x + \mu$, which is then used as input to the encoder model. Subsequently, the output of the encoder, denoted as $f_{e}(x')$, is utilized to extract latent features via a fully connected network. Additionally, $f_{e}(x')$ is passed as input to the decoder function to produce the output $f_{d}(f_{e}(x'))$. To evaluate the quality of the reconstruction, a mean square error is computed between $f_{d}(f_{e}(x'))$ and $x$, as shown in Eqn.\ref{eqn_4.5}.
\begin{gather}
    \text{Loss} = ||f_{d}(f_{e}(x')) - x||^{2}
\label{eqn_4.5}    
\end{gather}

The reconstruction loss measures how well the decoder can generate a reconstructed input from the encoded features. The 3D denoising autoencoder consists of three layers of 3D convolutional networks followed by batch normalization, Recitified Linear Unit (Relu) activation function, and 3D max pooling. The latent features extracted from the encoder's output are used for the fusion with genomics. The detailed architecture of the 3D convolution-based autoencoder is provided in Supplementary Table 1.

\subsection{Radiogenomic Bipartite Graph Construction}
\begin{figure*}[tb]
\centering
\small{\text{A}}\includegraphics[width= 1.7in]{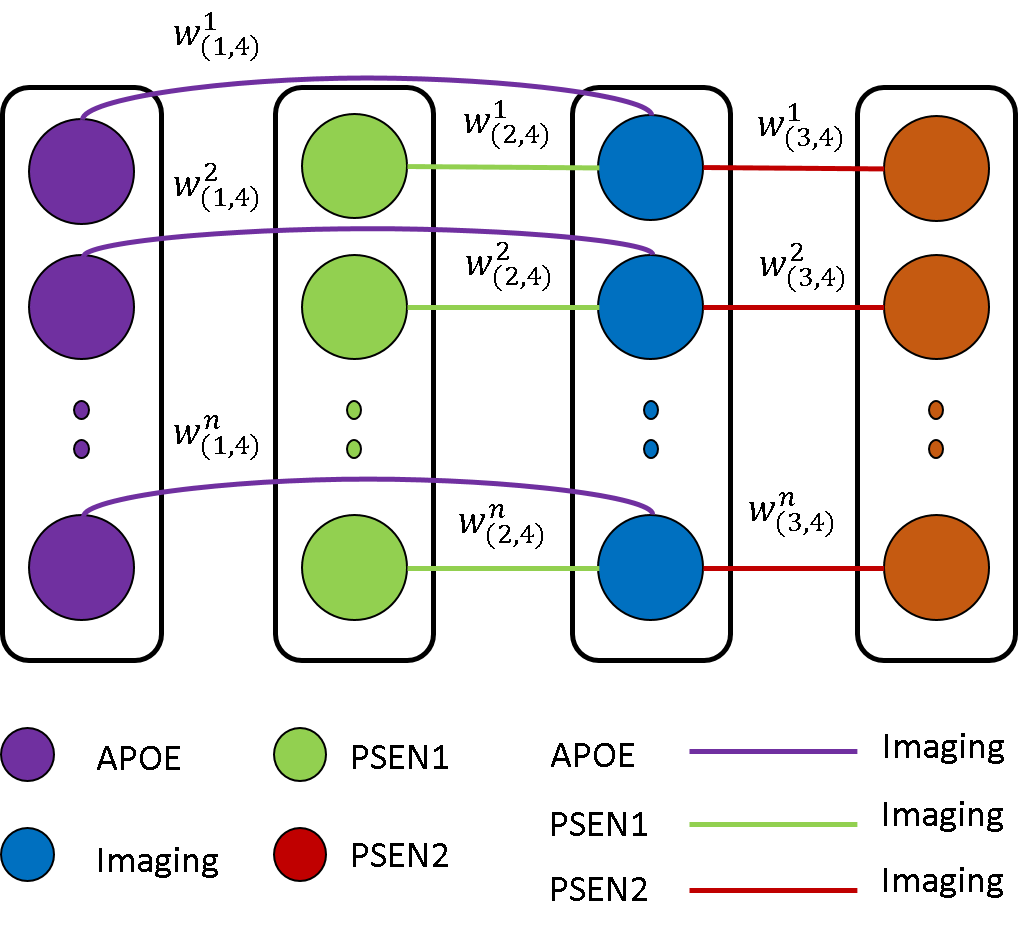}
\small{\text{B}}\includegraphics[width= 1.7in]{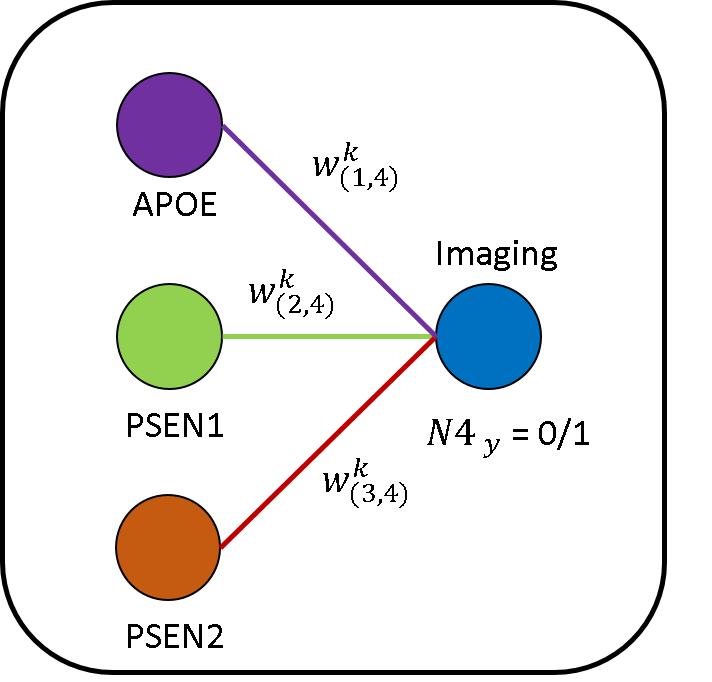}
\caption{\label{fig:fig_4.2} The bipartite graph construction for imaging-genomics fusion in AD detection. (A) graph containing all samples in the study. (B) Subgraph representing one sample.}
\end{figure*}

A bipartite graph $G=(U,V,E)$ is represented by two sets of different domains $U$(genes) and $V$(Images), and a set of edges $E \subset U \times V$. Bipartite graphs contain only inter-domain edges between genes and images. Each sample represents a subgraph with $|U|= 3$, representing three AD driver genes (APOE, PSEN1, and PSEN2), $|V|= 1$  representing the corresponding MRI image, and $|E|=3$ which represents three different types of edges: $\{\text{'Edge 1': } \text{Node 1} \rightarrow \text{Node 4}$, $\text{'Edge 2': } \text{Node 2} \rightarrow \text{Node 4}$, and $\text{'Edge 3': } \text{Node 3} \rightarrow \text{Node 4} \}$. Fig.~\ref{fig:fig_4.2}.A represents the bipartite graph structure for all the samples, while Fig.~\ref{fig:fig_4.2}.B illustrates one single subgraph representing one sample in the study. The label $N4_{y}$ is used to indicate whether the sample belongs to the AD class ($N4_{y} = 1$) or another class, such as MCI, in a binary classification ($N4_{y} = 0$).

\subsection{Bipartite Graph Neural Network}
A graph neural network (GNN) consists of three fundamental steps: message passing, aggregating, and updating \cite{zhou2020graph}. Consider a given graph $G = (V,E)$, where $V$ represents  the set of nodes, and $X \in R^{d \times |V|}$ represents the node features, where $d$ is dimension of each node feature, and $|V|$ is the total number of nodes in the graph. Let's assume the node embedding denoted as $h_{u}^{k}$ at step $k$,  where $u\in \{V \} $. According to the Update function, the node embedding is updated based on its neighborhood, represented as $N(u)$, as \cite{wu2022graph}:
\begin{gather}
    h_{u}^{(k+1)} = \text{Update}^{(k)}(h_{u}^{(k)}, m_{N(u)}^{(k)})
\label{eqn_4.2}
\end{gather}
where $h_{u}^{(k+1)}$ represents the updated node embedding at step $k+1$ and is calculated by applying the Update function to the previous embedding $h_{u}^{(k)}$ and the aggregated messages $m_{N(u)}^{(k)}$. The aggregation process involves collecting information from neighboring nodes (such as node $v$). We define these two functions with weights $w_{self}$ and $w_{neigh}$ as the following.
\begin{gather}
    \text{Update}(h_{u}, m_{N_{u}}) = \sigma(W_{self}(h_{u}) + W_{neigh}(m_{N_{u}}))
\label{eqn_4.3}    
\end{gather}
where $\sigma$ represent a non linearity. Finally, the message passing in GNN can be expressed as: 
\begin{gather}
    h_{u}^{(k+1)} = \sigma( W_{self}^{(k)}(h_{u}^{(k)}) + W^{(k)}_{neigh}(\sum_{v \in N(u)}h_{v}^{(k)})  + b)
\label{eqn_4.4}
\end{gather}
where $b$ denotes the bias term.
In the proposed approach, the goal is to learn a dynamic adjacency matrix that establishes connections between different types of nodes based on their respective node features.
During training, the weights associated with each edge are dynamically updated. The primary objective is to identify the optimal edge weights, denoted as $w_{(1,4)}$, $w_{(2,4)}$, and $w_{(3,4)}$, while classifying the subgraphs extracted from the bipartite graph into AD, MCI, or CN groups. We introduce an aggregation function for learning weights associated with each type of edge, which is derived from a randomly sampled prior. 
This approach stands in contrast to the graph attention network (GAT) (\cite{velivckovic2017graph}), where the attention parameters are implicitly learned through a neural network using concatenated node feature embeddings as input.
In our formulation, we have four types of nodes as $i,j,k$ and $l$; and four types of edges: $i\rightarrow l$, $j\rightarrow l$ and $k\rightarrow l$. 
Further, let  $|V_{i}|,$$|V_{j}|$, $|V_{k}|$ and $|V_{l}|$ represent the number of neighboring nodes around each type of node. The node features for each type are represented as $X_{i} \in R^{d_{i}\times |V_{i}|}$, $X_{j} \in R^{d_{j}\times |V_{j}|}$, $X_{k} \in R^{d_{k}\times |V_{k}|}$ and $X_{l} \in R^{d_{l}\times |V_{l}|}$.  
\begin{figure*}[tb]
\centering
\includegraphics[width=\textwidth]{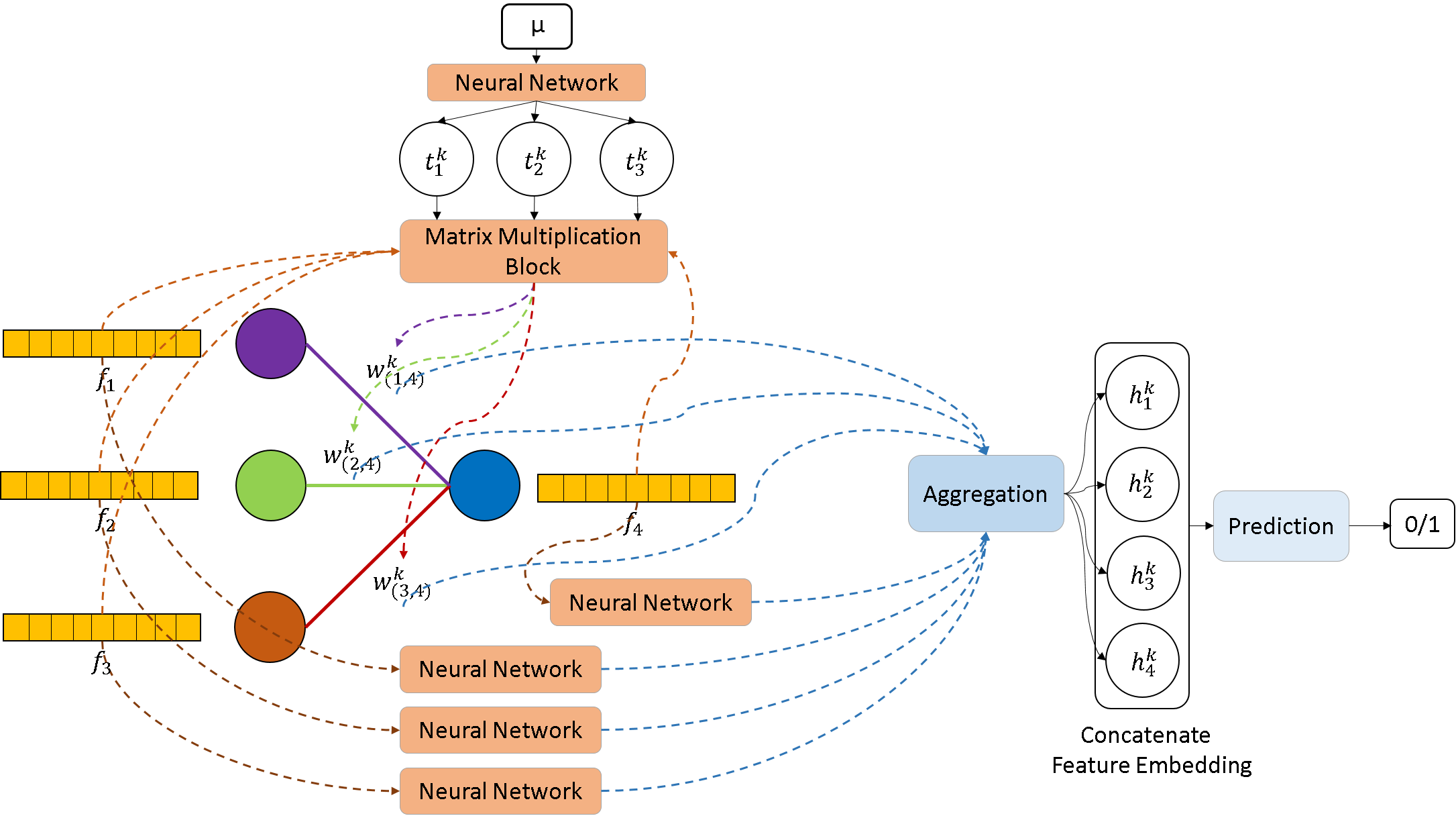}
\caption{\label{fig:framework} The heterogeneous bipartite GNN model with dynamic edge weight learning.}
\end{figure*}
We initialize a prior $\alpha$ sampled from a random Gaussian distribution with a mean of 0 and a standard variation of 1. This prior remains consistent for all types of nodes. Subsequently, we compute a higher dimensional representation $z$ for each type of edge using the formula $z = \phi \alpha $, where, $\phi$ represents a differentiable learnable function, in this case a fully connected neural network layer. Following this, we calculate the edge weights, denoted by $e$. For instance, the edge weights $e_{i,l}$ is computed as $e_{i,l} = \sigma (x_{i}.z.x_{l}^{T})$, where $\sigma$ is non-linearity(Relu). We then compute the node embedding, represented as $h$, for each node using the formula $h = Vx$, where $V$ is a neural network with learnable parameters. Finally, we employ the aggregation function to compute the updated embedding for each node, as:
\begin{gather}
    h_{p}^{n+1} = \sigma (\sum_{q \in N(p)} e_{p,q}^{n} h_{q}^{n})
\label{eqn_4.6}    
\end{gather}
where $n$ denotes the iteration, $p$ is the node for which we wish to compute the node embedding and $q$ represents the nodes in the neighborhood of $p$.

The framework of the proposed heterogeneous graph neural network is shown in Fig.\ref{fig:framework}, where $f_{i}$ denotes the features associated with each node, and $t_{j}$ represents the higher dimensional feature representation for each type of edge in the graph. The core component of the network includes a matrix multiplication block, which takes as input the feature vectors and $t_{j}$, resulting in the computation of edge weights $w_{j}$ associated with each edge. The calculated weights in junction with the feature embeddings, are subsequently supplied as input to the aggregation block. The output of this aggregation block serves as the input to the prediction block, which generates the final classification score. Throughout this process, the loss is computed and back-propagated through the neural network blocks.
Finally, we compute the averaged weight parameters for positive and negative samples per feature dimension in the test set to identify which gene contributes more significantly to the final graph classification score. Eqn.\ref{eqn:avergae_cal} outlines the formula employed to compute the average weights for both classes separately. In this equation,  $p$ can take values $\{0,1,2\}$ representing the three genes, $d_{p}$ represents the features dimension of the $p^{th}$ node, and $N$ corresponds to the total number of subgraphs.
%
 \begin{gather}
     w^{avg}_{(p,4)} = \frac{1}{d_{p}}\sum_{k = 1}^{N}w^{k}_{p,4}
 \label{eqn:avergae_cal}    
 \end{gather}

\section{Experiments and Results}
\textbf{Datasets}. All image and genomics data are publicly available. We collected MRI images and genomic data for three AD driver genes \cite{Lanoiselee2022APP}, namely APOE (Apolipoprotein E), PSEN1 (Presenilin 1), and PSEN2 (Presenilin 2) from the Alzheimer's Disease Neuroimaging Initiative (ADNI) $https://adni.loni.usc.edu/$. 
A total of $52$ samples, with an equal number of subjects representing three classes (CN, MCI, and AD), were collected from ADNI2 and ADNIGO datasets. Each 3D MRI image had dimension of $170\times 256 \times 256$ and underwent preprocessing using a histogram-based ranking method \cite{raj2022end}. This preprocessing step reduced the number of slices per sample in the 3D MRI to $64$ slices. These processed samples served as input for the denoising autoencoder for feature extraction. The final imaging features had dimension of $52\times 512$.
The gene expression data dimensions were as follows: $52\times 1$ for $APOE$, $52\times 4$ for $PSEN1$, and $52\times 4$ for $PSEN2$. The genomics features were normalized using the min-max normalization. 
Samples with missing corresponding genomics data were excluded from the dataset.
\newline
\newline
\textbf{Implementation Details}. The denoising autoencoder feature extraction was trained for total of $100$ epochs using Adam optimizer with a learning rate of $9e-4$. Additionally, a cosine annealing learning rate scheduler was employed. The bipartite GNN training was conducted over $800$ epochs with the Adam optimizer and learning rate of $9e-3$. Both the autoencoder and GNN models were trained using the mean square error loss function. Experiments were performed using all three genes for learning, as well as all possible permutations involving two genes. This allowed us to thoroughly investigate the effects of different gene combinations on our model's performance. We conducted binary classification experiments distinguishing between MCI vs. CN, MCI vs. AD, CN vs AD, and a three-way classification task involving MCI, AD, and CN. The feature dimensions of the genes and imaging data remained consistent across these classification tasks. The total number of samples used for training were 115, 52, and 78 for three additional classification tasks. In all cases, a standard 80-20 train-test split was employed. 
\newline
\newline
\textbf{Results.} In Table \ref{tab:tab_comparison}, our approach shows improved prediction accuracy by integrating the MRI image and the genomics data for the three genes. Our method outperforms state-of-the-art- models in all AD vs CN, AD vs MCI, CN vs MCI by integrating both image and genomics modality. Additionally, the the performance metrics for binary classification (MCI vs AD, MCI vs NC, AD vs NC) and 3-level classification (NC vs MCI vs AD) tasks using accoracy, F1-score, recall, precision is provided in Supplementary Table 2. The highest accuracy of $92\%$  and F1-score of $93\%$ is achieved for the classification of AD vs CN using the proposed weight learning. The model also achieved slightly less results for MCI and AD classification (accuracy of $91\%$ and F1 score of $92.3\%$).
Additionally, Supplementary Table 3 shows the averaged weights learned over test data. The computation of edge weights is carried out using the formula outlined in Eqn.\ref{eqn:avergae_cal}. 
%
\begin{table*}
\scriptsize
\centering
\scalebox{1}{
\begin{tabular}{|lll|lll|lll|lll|}
\hline
\multicolumn{3}{|l|}{\textbf{Classification Task :}} &
  \multicolumn{3}{l|}{\textbf{AD vs. CN \%}} &
  \multicolumn{3}{l|}{\textbf{AD vs. MCI \%}} &
  \multicolumn{3}{l|}{\textbf{CN vs. MCI \%}} \\ \hline
\multicolumn{1}{|l|}{\begin{tabular}[c]{@{}l@{}}  \\ \end{tabular}} &
  \multicolumn{1}{l|}{Method} &
  Modality &
  \multicolumn{1}{l|}{Acc} &
  \multicolumn{1}{l|}{F1} &
  BA &
  \multicolumn{1}{l|}{Acc} &
  \multicolumn{1}{l|}{F1} &
  BA &
  \multicolumn{1}{l|}{Acc} &
  \multicolumn{1}{l|}{F1} &
  BA \\ \hline
\multicolumn{1}{|l|}{\multirow{4}{*}{\cite{peng2016structured}}} &
  \multicolumn{1}{l|}{\multirow{4}{*}{SSR}} &
  MRI &
  \multicolumn{1}{l|}{88.4} &
  \multicolumn{1}{l|}{88.32} &
  88.5 &
  \multicolumn{1}{l|}{-} &
  \multicolumn{1}{l|}{-} &
  - &
  \multicolumn{1}{l|}{71.6} &
  \multicolumn{1}{l|}{60.41} &
  65.5 \\ \cline{3-12} 
\multicolumn{1}{|l|}{} &
  \multicolumn{1}{l|}{} &
  \begin{tabular}[c]{@{}l@{}}SNP  (135 genes)\end{tabular} &
  \multicolumn{1}{l|}{76} &
  \multicolumn{1}{l|}{75.66} &
  76.2 &
  \multicolumn{1}{l|}{-} &
  \multicolumn{1}{l|}{-} &
  - &
  \multicolumn{1}{l|}{66.2} &
  \multicolumn{1}{l|}{58.73} &
  61.75 \\ \cline{3-12} 
\multicolumn{1}{|l|}{} &
  \multicolumn{1}{l|}{} &
  \begin{tabular}[c]{@{}l@{}}MRI + SNP\end{tabular} &
  \multicolumn{1}{l|}{91.1} &
  \multicolumn{1}{l|}{91.17} &
  91.2 &
  \multicolumn{1}{l|}{-} &
  \multicolumn{1}{l|}{-} &
  -&
  \multicolumn{1}{l|}{74.9} &
  \multicolumn{1}{l|}{67.14} &
  70.1 \\ \cline{3-12} 
\multicolumn{1}{|l|}{} &
  \multicolumn{1}{l|}{} &
  \begin{tabular}[c]{@{}l@{}}MRI + \\ PET + SNP\end{tabular} &
  \multicolumn{1}{l|}{96.1} &
  \multicolumn{1}{l|}{96.08} &
  96.1 &
  \multicolumn{1}{l|}{-} &
  \multicolumn{1}{l|}{-} &
  - &
  \multicolumn{1}{l|}{80.3} &
  \multicolumn{1}{l|}{76.89} &
  77.7 \\ \hline
\multicolumn{1}{|l|}{\multirow{4}{*}{\cite{zhang2014integrative}}} &
  \multicolumn{1}{l|}{\multirow{4}{*}{SMML}} &
  \begin{tabular}[c]{@{}l@{}}SNP (189 genes)\end{tabular} &
  \multicolumn{1}{l|}{62} &
  \multicolumn{1}{l|}{-} &
  - &
  \multicolumn{1}{l|}{64} &
  \multicolumn{1}{l|}{-} &
  - &
  \multicolumn{1}{l|}{66} &
  \multicolumn{1}{l|}{-} &
  - \\ \cline{3-12} 
\multicolumn{1}{|l|}{} &
  \multicolumn{1}{l|}{} &
  MRI &
  \multicolumn{1}{l|}{86} &
  \multicolumn{1}{l|}{-} &
  - &
  \multicolumn{1}{l|}{67} &
  \multicolumn{1}{l|}{-} &
  - &
  \multicolumn{1}{l|}{69} &
  \multicolumn{1}{l|}{-} &
  - \\ \cline{3-12} 
\multicolumn{1}{|l|}{} &
  \multicolumn{1}{l|}{} &
  \begin{tabular}[c]{@{}l@{}}MRI + SNP\end{tabular} &
  \multicolumn{1}{l|}{87} &
  \multicolumn{1}{l|}{-} &
  - &
  \multicolumn{1}{l|}{69} &
  \multicolumn{1}{l|}{\textbf{-}} &
  - &
  \multicolumn{1}{l|}{70} &
  \multicolumn{1}{l|}{-} &
  - \\ \cline{3-12} 
\multicolumn{1}{|l|}{} &
  \multicolumn{1}{l|}{} &
  \begin{tabular}[c]{@{}l@{}}MRI + PET\\ + CSF + SNP\end{tabular} &
  \multicolumn{1}{l|}{94.8} &
  \multicolumn{1}{l|}{94.74} &
  94.75 &
  \multicolumn{1}{l|}{69.1} &
  \multicolumn{1}{l|}{69.24} &
  69.25 &
  \multicolumn{1}{l|}{75.6} &
  \multicolumn{1}{l|}{74.87} &
  74.9 \\ \hline
\multicolumn{1}{|l|}{\multirow{2}{*}{\cite{yan2017identification}}} &
  \multicolumn{1}{l|}{DSCCA} &
  \multirow{2}{*}{\begin{tabular}[c]{@{}l@{}}MRI + CSF \\ + Plasma \end{tabular}} &
  \multicolumn{1}{l|}{92.12} &
  \multicolumn{1}{l|}{-} &
  - &
  \multicolumn{1}{l|}{70.33} &
  \multicolumn{1}{l|}{-} &
  - &
  \multicolumn{1}{l|}{75.26} &
  \multicolumn{1}{l|}{-} &
  - \\ \cline{2-2} \cline{4-12} 
\multicolumn{1}{|l|}{} &
  \multicolumn{1}{l|}{PMA} &
   &
  \multicolumn{1}{l|}{83.38} &
  \multicolumn{1}{l|}{-} &
  - &
  \multicolumn{1}{l|}{67.72} &
  \multicolumn{1}{l|}{-} &
  - &
  \multicolumn{1}{l|}{53.51} &
  \multicolumn{1}{l|}{-} &
  - \\ \hline
\multicolumn{1}{|l|}{\multirow{4}{*}{\cite{maddalena2023integrating}}} &
  \multicolumn{1}{l|}{\multirow{3}{*}{\begin{tabular}[c]{@{}l@{}} \\ \end{tabular}}} &
  \begin{tabular}[c]{@{}l@{}}SNP \\ \end{tabular} &
  \multicolumn{1}{l|}{85.3} &
  \multicolumn{1}{l|}{47.6} &
  69.6 &
  \multicolumn{1}{l|}{88.7} &
  \multicolumn{1}{l|}{35.3} &
  64.9 &
  \multicolumn{1}{l|}{59.1} &
  \multicolumn{1}{l|}{71.9} &
  49.8 \\ \cline{3-12} 
\multicolumn{1}{|l|}{} &
  \multicolumn{1}{l|}{} &
  MRI &
  \multicolumn{1}{l|}{92.2} &
  \multicolumn{1}{l|}{68.1} &
  79 &
  \multicolumn{1}{l|}{88.4} &
  \multicolumn{1}{l|}{26.4} &
  59.5 &
  \multicolumn{1}{l|}{63.2} &
  \multicolumn{1}{l|}{71.5} &
  59.7 \\ \cline{3-12} 
\multicolumn{1}{|l|}{} &
  \multicolumn{1}{l|}{} &
  MRI + SNP &
  \multicolumn{1}{l|}{94.1} &
  \multicolumn{1}{l|}{76.4} &
  83.6 &
  \multicolumn{1}{l|}{91.8} &
  \multicolumn{1}{l|}{48.7} &
  70.9 &
  \multicolumn{1}{l|}{61.7} &
  \multicolumn{1}{l|}{71.2} &
  56.8 \\ \cline{1-12} 
\multicolumn{1}{|l|}{\cite{bi2020multimodal}} &
  \multicolumn{1}{l|}{\begin{tabular}[c]{@{}l@{}}CERF \\ + SVM\end{tabular}} &
  \begin{tabular}[c]{@{}l@{}}fMRI + SNP\end{tabular} &
  \multicolumn{1}{l|}{87} &
  \multicolumn{1}{l|}{-} &
  - &
  \multicolumn{1}{l|}{\textbf{-}} &
  \multicolumn{1}{l|}{-} &
  - &
  \multicolumn{1}{l|}{80} &
  \multicolumn{1}{l|}{-} &
  - \\ \hline
\multicolumn{1}{|l|}{\cite{raj2022multi}} &
  \multicolumn{1}{l|}{\begin{tabular}[c]{@{}l@{}}RL + \\ reg. \end{tabular}} &
  \begin{tabular}[c]{@{}l@{}}GE (3 genes)\end{tabular} &
  \multicolumn{1}{l|}{70.31} &
  \multicolumn{1}{l|}{70} &
  70 &
  \multicolumn{1}{l|}{63.80} &
  \multicolumn{1}{l|}{61.75} &
  62.30 &
  \multicolumn{1}{l|}{58.33} &
  \multicolumn{1}{l|}{57.77} &
  57.87 \\ \cline{2-12}
  \hline
\multicolumn{1}{|l|}{\cite{raj2022end}} &
  \multicolumn{1}{l|}{LRCN} &
  MRI &
  \multicolumn{1}{l|}{71.28} &
  \multicolumn{1}{l|}{68.4} &
  69 &
  \multicolumn{1}{l|}{-} &
  \multicolumn{1}{l|}{-} &
  - &
  \multicolumn{1}{l|}{-} &
  \multicolumn{1}{l|}{-} &
  - \\ \cline{2-12}
  \hline
\multicolumn{1}{|l|}{\textbf{Ours}} &
  \multicolumn{1}{l|}{\textbf{BGRL}} &
  \textbf{MRI + GE} &
  \multicolumn{1}{l|}{\textbf{92}} &
  \multicolumn{1}{l|}{\textbf{93}} &
  \textbf{93.75} &
  \multicolumn{1}{l|}{\textbf{91}} &
  \multicolumn{1}{l|}{\textbf{92.3}} &
  \textbf{92.85} &
  \multicolumn{1}{l|}{\textbf{78}} &
  \multicolumn{1}{l|}{\textbf{78.26}} &
  \textbf{82.14} \\ \hline
\end{tabular}}
\caption{Comparison with existing works for the task of AD vs. CN, AD vs. MCI and CN vs. MCI binary classification using ADNI dataset}
\label{tab:tab_comparison}
\normalsize
\end{table*}

\section{Ablation Study}
We verified the model efficiency by evaluating the effect of graph structure learning. Specifically, we trained the model with and without dynamic edge weights, as shown in Table \ref{tab:tab_4.2}. It can be observed that the model achieves an accuracy of $92\%$, utilizing the learned edge weights, which represents a $17\%$ improvement compared to the scenario without using weights.
\begin{table*}[tb]
\centering
\scalebox{1}{
\begin{tabular}{|lllllll|}
\hline
\multicolumn{7}{|l|}{Experiment with all four nodes : Imaging $+$ Genes (APOE, PSEN1, PSEN2)} \\ \hline
\multicolumn{1}{|l|}{\begin{tabular}[c]{@{}l@{}}Learning \\ weights\end{tabular}} &
\multicolumn{1}{|l|}{\begin{tabular}[c]{@{}l@{}}F1 \\ score\end{tabular}} &
  \multicolumn{1}{l|}{Recall} &
  \multicolumn{1}{l|}{Precision} &
  \multicolumn{1}{l|}{Accuracy} &
  \multicolumn{1}{|l|}{\begin{tabular}[c]{@{}l@{}}Macro \\ Average\end{tabular}} &
  \begin{tabular}[c]{@{}l@{}}Weighted \\ Average\end{tabular} \\ \hline
\multicolumn{1}{|l|}{w/o} &
  \multicolumn{1}{l|}{60\%} &
  \multicolumn{1}{l|}{100\%} &
  \multicolumn{1}{l|}{42\%} &
  \multicolumn{1}{l|}{75\%} &
  \multicolumn{1}{l|}{70\%} &
  72\% \\ \hline
\multicolumn{1}{|l|}{w} &
  \multicolumn{1}{l|}{\textbf{93\%}} &
  \multicolumn{1}{l|}{\textbf{100\%}} &
  \multicolumn{1}{l|}{\textbf{87.5\%}} &
  \multicolumn{1}{l|}{\textbf{92\%}} &
  \multicolumn{1}{l|}{\textbf{91\%}} &
  \textbf{91\%} \\ \hline
\end{tabular}}
\caption{Performance of the proposed model for the AD vs. CN classification task on the test set, comparing results with and without dynamic weight learning using all three genes for training }
\label{tab:tab_4.2}
\end{table*}
Additionally, we assessed the effect of the integrated model by retaining two genes and imaaging while removing one gene. The classification results based on two gene permutations, with and without dynamic weight learning, are shown in Supplementary Table 4. Among the gene permutations, the combination of $\{PSEN1$, $PSEN2, MRI\}$ provided the most robust performance across all metrics. With learned weights, this set achieved an accuracy of $83.33\%$, surpassing the performance without weights by $16\%$. Nevertheless, the performance of the two genes and the image remains inferior to the accuracy achieved with all three genes and image combined. In fact, the integrated performance attained an accuracy of $92\%$, F1 score of $93\%$, recall of $100\%$, and precision of $87.5\%$ when incorporating all three genes and the image, surpassing the performance achieved by omitting any individual modality.

\section{Conclusion}
In this study, we proposed a novel multimodal bipartite graph representation learning to exploit the complementary relationship of genomics and MRI images for Alzheimer's disease detection. Specifically, we introduced a fusion model (gene expression and structural MRI) using a bipartite graph in which the edges are leaned dynamically through the training of a graph neural network. Additionally, we identified the importance of the edges by introducing averaged weight over the edges. Experimental results demonstrate the superior performance of the proposed approach compared to the state-of-the-art methods.
\newline  
\textbf{Acknowledgment.}
The results of this study are based on the data collected from the public ADNI dataset $https://adni.loni.usc.edu/$. 





\begin{thebibliography}{10}

\bibitem{agartz1999hippocampal}
Ingrid Agartz, Reza Momenan, Robert~R Rawlings, Michael~J Kerich, and Daniel~W
  Hommer.
\newblock Hippocampal volume in patients with alcohol dependence.
\newblock {\em Archives of general psychiatry}, 56(4):356--363, 1999.

\bibitem{bi2020multimodal}
Xia-an Bi, Xi~Hu, Hao Wu, and Yang Wang.
\newblock Multimodal data analysis of alzheimer's disease based on clustering
  evolutionary random forest.
\newblock {\em IEEE Journal of Biomedical and Health Informatics},
  24(10):2973--2983, 2020.

\bibitem{camicioli2003parkinson}
Richard Camicioli, M~Milar Moore, Anthony Kinney, Elizabeth Corbridge, Kathryn
  Glassberg, and Jeffrey~A Kaye.
\newblock Parkinson's disease is associated with hippocampal atrophy.
\newblock {\em Movement disorders}, 18(7):784--790, 2003.

\bibitem{chaddad2018deep}
Ahmad Chaddad, Christian Desrosiers, and Tamim Niazi.
\newblock Deep radiomic analysis of mri related to alzheimer’s disease.
\newblock {\em Ieee Access}, 6:58213--58221, 2018.

\bibitem{fujioka2000hippocampal}
Masayuki Fujioka, Kenji Nishio, Seiji Miyamoto, Ken-Ichiro Hiramatsu, Toshisuke
  Sakaki, Kazuo Okuchi, Toshiaki Taoka, and Susumu Fujioka.
\newblock Hippocampal damage in the human brain after cardiac arrest.
\newblock {\em Cerebrovascular diseases}, 10(1):2--7, 2000.

\bibitem{Gupta2019Prediction}
Yabraj Gupta, Ramesh~Kumar Lama, Goo-Rak Known, and the Alzheimer's Disease
  Neuroimaging~Initiative.
\newblock Prediction and classification of alzheimer’s disease based on
  combined features from apolipoprotein-e genotype, cerebrospinal fluid, mr,
  and fdg-pet imaging biomarkers.
\newblock In {\em Frontiers in Computational Neuroscience}, volume~13. IEEE,
  2019.

\bibitem{jiang2022using}
Jiehui Jiang, Min Wang, Ian Alberts, Xiaoming Sun, Taoran Li, Axel Rominger,
  Chuantao Zuo, Ying Han, Kuangyu Shi, and for the Alzheimer’s
  Disease~Neuroimaging Initiative.
\newblock Using radiomics-based modelling to predict individual progression
  from mild cognitive impairment to alzheimer’s disease.
\newblock {\em European journal of nuclear medicine and molecular imaging},
  49(7):2163--2173, 2022.

\bibitem{johnson2012brain}
Keith~A Johnson, Nick~C Fox, Reisa~A Sperling, and William~E Klunk.
\newblock Brain imaging in alzheimer disease.
\newblock {\em Cold Spring Harbor perspectives in medicine}, 2(4):a006213,
  2012.

\bibitem{Kavitha2022Early}
C~Kavitha, Vinodhini mani, SR~Srividhya, Osamah~Ibrahim Khalaf, and Carlos
  Andres~Tavera Romero.
\newblock Early-stage alzheimer's disease prediction using machine learning
  models.
\newblock In {\em Frontiers in Public Health}, 2022.

\bibitem{Lanoiselee2022APP}
Helen-Marie Lanoiselee, Gael Nicolas, David Wallon, Anne Rovelet-Lecrux,
  Morgane Lacour, and et~al.
\newblock App, psen1, and psen2 mutations in early-onset alzheimer disease: A
  genetic screening study of familial and sporadic cases.
\newblock In {\em PLOS Medicine}, volume~4, 2017.

\bibitem{li2020radiomics}
Tao-Ran Li, Yue Wu, Juan-Juan Jiang, Hua Lin, Chun-Lei Han, Jie-Hui Jiang, and
  Ying Han.
\newblock Radiomics analysis of magnetic resonance imaging facilitates the
  identification of preclinical alzheimer’s disease: an exploratory study.
\newblock {\em Frontiers in Cell and Developmental Biology}, 8:605734, 2020.

\bibitem{maddalena2023integrating}
Lucia Maddalena, Ilaria Granata, Maurizio Giordano, Mario Manzo, and
  Mario~Rosario Guarracino.
\newblock Integrating different data modalities for the classification of
  alzheimer’s disease stages.
\newblock {\em SN Computer Science}, 4(3):249, 2023.

\bibitem{Mirzaei2021Ensemble}
Golrokh Mirzaei.
\newblock Ensembles of convolutional neural network pipelines for diagnosis of
  alzheimer’s disease.
\newblock In {\em IEEE 2021 55th Asilomar Conference on Signals, Systems, and
  Computers}, pages 583--589. IEEE, 2021.

\bibitem{mirzaei2016ImagingMachineLearning}
Golrokh Mirzaei, Anahita Adeli, and Hojjat Adeli.
\newblock Imaging and machine learning techniques for diagnosis of
  alzheimer’s disease.
\newblock In {\em Reviews in the Neurosciences}, volume~27, pages 857--870,
  2016.

\bibitem{mirzaei2022MachineLearning}
Golrokh Mirzaei and Hojjat Adeli.
\newblock Machine learning techniques for diagnosis of alzheimer disease, mild
  cognitive disorder, and other types of dementia.
\newblock In {\em Biomedical Signal Processing and Control}, volume~72.
  Elsevier, 2022.

\bibitem{Nativio2020An}
Raffaella Nativio, Yemin Lan, Greg Donahue, and et~al.
\newblock An integrated multi-omics approach identifies epigenetic alterations
  associated with alzheimer's disease.
\newblock In {\em Nature Genetics}, 2020.

\bibitem{peng2016structured}
Jailin Peng, Le~An, Xiaofeng Zhu, Yan Jin, and Dinggang Shen.
\newblock Structured sparse kernel learning for imaging genetics based
  alzheimer’s disease diagnosis.
\newblock In {\em Medical Image Computing and Computer-Assisted
  Intervention--MICCAI 2016: 19th International Conference, Athens, Greece,
  October 17-21, 2016, Proceedings, Part II 19}, pages 70--78. Springer, 2016.

\bibitem{raj2022end}
Aditya Raj and Golrokh Mirzaei.
\newblock End to end trained long term recurrent convolutional network for
  subject-level alzheimer detection.
\newblock In {\em 2022 56th Asilomar Conference on Signals, Systems, and
  Computers}, pages 303--306. IEEE, 2022.

\bibitem{raj2022multi}
Aditya Raj and Golrokh Mirzaei.
\newblock Multi-armed bandit approach for multi-omics integration.
\newblock In {\em 2022 IEEE International Conference on Bioinformatics and
  Biomedicine (BIBM)}, pages 3130--3136. IEEE, 2022.

\bibitem{rosas2003evidence}
Herminia~D Rosas, WJ~Koroshetz, YI~Chen, C~Skeuse, M~Vangel, ME~Cudkowicz,
  K~Caplan, K~Marek, LJ~Seidman, N~Makris, et~al.
\newblock Evidence for more widespread cerebral pathology in early hd: an
  mri-based morphometric analysis.
\newblock {\em Neurology}, 60(10):1615--1620, 2003.

\bibitem{salvatore2021radiomics}
Christian Salvatore, Isabella Castiglioni, and Antonio Cerasa.
\newblock Radiomics approach in the neurodegenerative brain.
\newblock {\em Aging Clinical and Experimental Research}, 33:1709--1711, 2021.

\bibitem{van2021imaging}
Wieke~M van Oostveen and Elizabeth~CM de~Lange.
\newblock Imaging techniques in alzheimer’s disease: a review of applications
  in early diagnosis and longitudinal monitoring.
\newblock {\em International journal of molecular sciences}, 22(4):2110, 2021.

\bibitem{velivckovic2017graph}
Petar Veli{\v{c}}kovi{\'c}, Guillem Cucurull, Arantxa Casanova, Adriana Romero,
  Pietro Lio, and Yoshua Bengio.
\newblock Graph attention networks.
\newblock {\em arXiv preprint arXiv:1710.10903}, 2017.

\bibitem{wu2022graph}
Lingfei Wu, Peng Cui, Jian Pei, Liang Zhao, and Le~Song.
\newblock {\em Graph neural networks}.
\newblock Springer, 2022.

\bibitem{yan2017identification}
Jingwen Yan, Shannon~L Risacher, Kwangsik Nho, Andrew~J Saykin, and LI~Shen.
\newblock Identification of discriminative imaging proteomics associations in
  alzheimer's disease via a novel sparse correlation model.
\newblock In {\em PSB}, pages 94--104. World Scientific, 2017.

\bibitem{zhang2014integrative}
Ziming Zhang, Heng Huang, Dinggang Shen, and Alzheimer's Disease~Neuroimaging
  Initiative.
\newblock Integrative analysis of multi-dimensional imaging genomics data for
  alzheimer's disease prediction.
\newblock {\em Frontiers in aging neuroscience}, 6:260, 2014.

\bibitem{zhou2019dual}
Hucheng Zhou, Jiehui Jiang, Jiaying Lu, Min Wang, Huiwei Zhang, Chuantao Zuo,
  and Alzheimer’s Disease~Neuroimaging Initiative.
\newblock Dual-model radiomic biomarkers predict development of mild cognitive
  impairment progression to alzheimer’s disease.
\newblock {\em Frontiers in neuroscience}, 12:1045, 2019.

\bibitem{zhou2020graph}
Jie Zhou, Ganqu Cui, Shengding Hu, Zhengyan Zhang, Cheng Yang, Zhiyuan Liu,
  Lifeng Wang, Changcheng Li, and Maosong Sun.
\newblock Graph neural networks: A review of methods and applications.
\newblock {\em AI open}, 1:57--81, 2020.

\end{thebibliography}


\end{document}